\PassOptionsToPackage{svgnames,table}{xcolor}
\documentclass[11pt,letterpaper,twocolumn]{template/style}

\usepackage[numbers]{natbib}
\setcitestyle{square}
\usepackage{subcaption}
\usepackage{mathtools}
\usepackage{multirow}
\usepackage{algorithm}
\usepackage[noend]{algpseudocode}
\usepackage{cleveref}
\usepackage{xspace}

\newtheorem{proposition}{Proposition}
\AtBeginDocument{\crefname{proposition}{Proposition}{Propositions}\Crefname{proposition}{Proposition}{Propositions}}

\newcommand{\dinov}{\mbox{DINOv3}}
\newcommand{\ours}{Puzzle\xspace}
\newcommand{\eg}{\emph{e.g.}}
\newsavebox{\fitbox}
\newcommand{\fitwidth}[1]{\sbox{\fitbox}{#1}\ifdim\wd\fitbox>\linewidth\resizebox{\linewidth}{!}{\usebox{\fitbox}}\else\usebox{\fitbox}\fi}

\definecolor{cBlue}{HTML}{2F6FB0}
\definecolor{cOrange}{HTML}{D97B1E}
\definecolor{cRed}{HTML}{B23A3A}
\definecolor{cGreen}{HTML}{1E9A78}
\definecolor{cPurple}{HTML}{7B52A6}
\definecolor{gaingreen}{RGB}{0,128,64}
\definecolor{lossred}{RGB}{185,40,40}

\newcommand{\venue}[1]{{\scriptsize\color{black!70}#1}}

\newcommand{\cls}{\texttt{[CLS]}\xspace}

\DeclareMathOperator{\mean}{Avg.}
\DeclareMathOperator{\clip}{clip}

\newcommand{\gain}[1]{{\color{gaingreen}#1}}
\newcommand{\loss}[1]{{\color{lossred}#1}}
\newcommand{\gp}[1]{\,{\scriptsize\textcolor{gaingreen}{(+#1)}}}

\newcommand{\gz}[1]{\,{\scriptsize\textcolor{black!55}{(0.0)}}}

\newcommand{\gzt}[1]{\,{\tiny\textcolor{black!55}{(0.0)}}}
\newcommand{\srcmark}{--}

\renewcommand{\cls}{\textnormal{\texttt{[CLS]}}\xspace}
\newcommand{\affmark}[1]{\textsuperscript{\textcolor{BgPrimary600}{\bfseries #1}}}

\author{
  Wensong Xu\affmark{*}, Siqi Lu\affmark{*}, Jibao Yuan\affmark{*}, Yongbin Zheng\affmark{\ensuremath{\dagger}}\\
  \affmark{1}National University of Defense Technology\\
  {\footnotesize \affmark{*}Equal contribution.\quad \affmark{\ensuremath{\dagger}}Corresponding author.}\\
  {\footnotesize\texttt{xws\_97@nudt.edu.cn; lusiqi9803@163.com; yuanjibao2025@nudt.edu.cn; zybnudt@nudt.edu.cn}}
}

\correspondingauthor{Yongbin Zheng (\texttt{zybnudt@nudt.edu.cn})}

\correspondingauthor{Yongbin Zheng (\textnormal{\texttt{zybnudt@nudt.edu.cn}})}

\title{\texorpdfstring{\raisebox{-0.2em}{\includegraphics[height=1em]{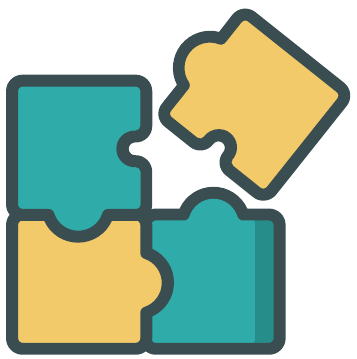}}\,Puzzle: One Pass Is Not Enough for Open-Vocabulary Remote Sensing Segmentation}{Puzzle: One Pass Is Not Enough for Open-Vocabulary Remote Sensing Segmentation}}
\runningtitle{Puzzle: Iterative Re-inference for OVRSIS}

\hypersetup{
  pdftitle={Puzzle: One Pass Is Not Enough for Open-Vocabulary Remote Sensing Segmentation},
  pdfauthor={Wensong Xu; Siqi Lu; Jibao Yuan; Yongbin Zheng},
  pdfkeywords={Remote Sensing Image Segmentation; Open-Vocabulary Semantic Segmentation}
}

\begin{document}
\raggedbottom

\begin{abstract}
Open-vocabulary remote sensing image segmentation (OVRSIS) enables the segmentation of unseen classes without additional annotation or retraining. By analyzing the predictions of existing OVRSIS methods, we find that many misclassified patches coexist with visually similar, correctly classified patches of the same class within the same image. Unfortunately, most of these methods stop after a single inference rather than using their predictions to guide further inference. Motivated by this observation, we propose \textbf{Puzzle}, a plug-and-play iterative re-inference framework that combines prediction history with a visual-similarity prior to progressively refine patch-level attention. Puzzle employs a lightweight history-conditioned controller to regulate each patch's contribution as an information source and its reliance on visually similar patches, while an adaptive stopping rule determines when to stop. It keeps the segmentation model frozen and reuses the same controller across different models and datasets without retraining. Extensive experiments demonstrate consistent performance improvements across diverse OVRSIS models and datasets, highlighting the strong transferability of Puzzle. Moreover, Puzzle can further improve the performance of SOTA methods.
\end{abstract}

\maketitle
\FloatBarrier

\section{Introduction}
\label{sec:intro}

Open-Vocabulary Remote Sensing Image Segmentation (OVRSIS) \cite{li2025segearthov,zhou2022maskclip} segments remote sensing (RS) images according to text-specified class names, including classes outside the training label set. Using vision-language models such as CLIP~\cite{radford2021clip}, OVRSIS can handle different class sets without additional annotations or model retraining.

\begin{figure*}[t]
\centering
\includegraphics[width=\linewidth]{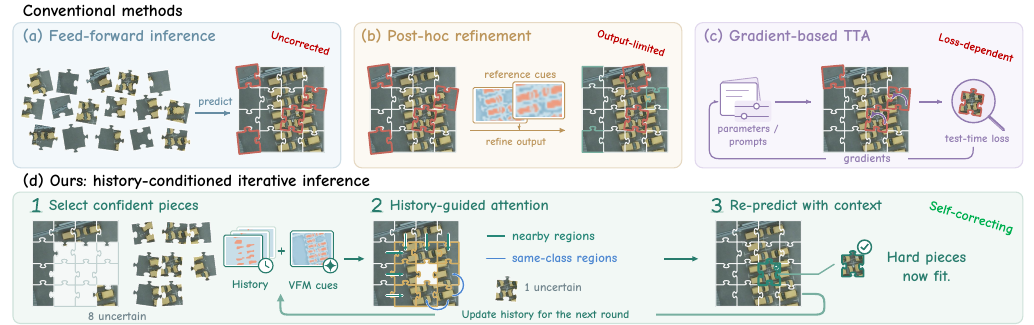}
\caption{Inference paradigms for open-vocabulary segmentation. Each jigsaw piece denotes an image patch, and red pieces denote misclassified patches. (a)--(c) Existing paradigms predict once, refine the output, or update parameters or prompts with a test-time loss. (d) Puzzle re-infers the frozen model and lets reliable pieces guide uncertain ones through attention.}
\label{fig:teaser}
\end{figure*}

\begin{figure}[t]
\centering
\includegraphics[width=\linewidth]{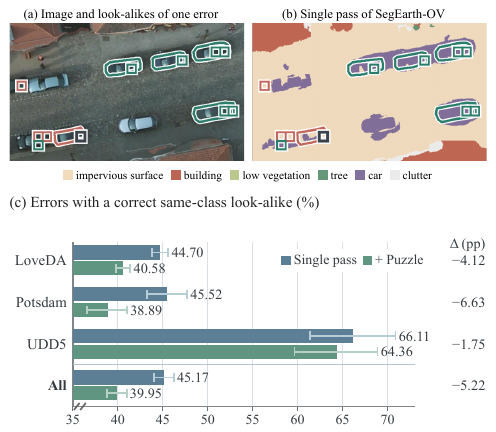}
\caption{Correctly classified look-alikes of misclassified patches. (a,b) A misclassified car patch (dark square) and its top-10 look-alikes (green: correct; red: misclassified) in a Potsdam image and the single-pass prediction of SegEarth-OV. (c) Share of misclassified patches with a correctly classified look-alike of the same class before and after applying Puzzle.}
\label{fig:evidence}
\end{figure}

Existing OVRSIS methods can be broadly categorized into three paradigms (Fig.~\ref{fig:teaser}(a)--(c)): single-pass methods modify visual representations or attention for dense prediction~\cite{lan2024clearclip,li2025segearthov,niu2026reattnclip,wang2024sclip,zhou2022maskclip}; post-hoc refinement methods use structural priors from vision foundation models (VFMs) to refine predictions~\cite{chen2026coninfer,stojnic2025lposs,wang2026sdci}; and test-time adaptation (TTA) methods update model parameters or prompts using unsupervised objectives~\cite{noori2025mlmp,shu2022tpt,wang2021tent,yang2026tmpa}. While effective, these methods can still produce inconsistent predictions across visually similar regions of the same class. Such regions are common in RS imagery, such as buildings within a block, ships in a row, and neighboring farmland parcels, where some are correctly classified while others are misclassified.

To examine how often this pattern occurs, we analyze the predictions of SegEarth-OV \cite{li2025segearthov} on LoveDA, Potsdam, and UDD5. As shown in Fig.~\ref{fig:evidence}, 45.2\% of misclassified patches have a visually similar, correctly classified region of the same class within the same image. This finding suggests that existing predictions may provide useful references for correction. This raises a question: \textbf{can these references guide the model to correct other visually similar regions in subsequent inference?}

A natural idea is to enable an existing OVRSIS model to re-infer over multiple rounds by adjusting information exchange among patches based on its prediction history and a VFM's visual-similarity prior, while keeping the model parameters fixed (Fig.~\ref{fig:teaser}(d)). However, this presents three challenges: (1) \textbf{Reference selection}: visually similar patches may also be misclassified, making reliable references difficult to identify. (2) \textbf{Reliance strength}: even when reliable references are available, different patches may need to rely on them to different degrees. (3) \textbf{Stopping point}: different images may require different numbers of rounds, while excessive re-inference may overwrite correct predictions.

To address these challenges, we propose \textbf{Puzzle}, a plug-and-play iterative re-inference method for OVRSIS. Like assembling a jigsaw puzzle, Puzzle uses more reliable regions from previous predictions to guide the correction of other visually similar regions over successive rounds. In Puzzle, a history-conditioned controller, trained once and reused across models and datasets, estimates each patch's source trust from the prediction history and a VFM's visual-similarity prior to give more weight to reliable references. The controller also assigns each patch a listening strength to control how much information it receives from similar patches, with both scores used to adjust the visual encoder's final-layer attention for re-inference. After each round, Puzzle evaluates prediction gain, rewriting risk, and changes between rounds to decide whether to continue, reducing the risk of overwriting correct predictions. Puzzle reduces this proportion by 5.2 percentage points overall (Fig.~\ref{fig:evidence}(c)), indicating that it makes effective use of the correction potential in existing predictions. Furthermore, experiments show that Puzzle improves segmentation accuracy across all 84 cross-dataset and cross-model evaluations and outperforms existing SOTA methods when applied to TMPA \cite{yang2026tmpa}.

Our contributions are summarized as follows:

\begin{itemize}
  \item We identify an inconsistency in OVRSIS predictions, where visually similar regions of the same class can be partly classified correctly and partly misclassified.
  
  \item We propose Puzzle, a plug-and-play iterative re-inference method that uses prediction history and a VFM's visual-similarity prior to regulate patch-level information exchange and adaptively determines when to stop, while keeping the underlying segmentation model frozen.
  
  \item Extensive experiments demonstrate that Puzzle consistently improves segmentation performance across different models and datasets and outperforms existing SOTA methods when applied to TMPA.
\end{itemize}

\section{Preliminaries}
\label{sec:prelim}

\begin{figure*}[t]
\centering
\includegraphics[width=\linewidth]{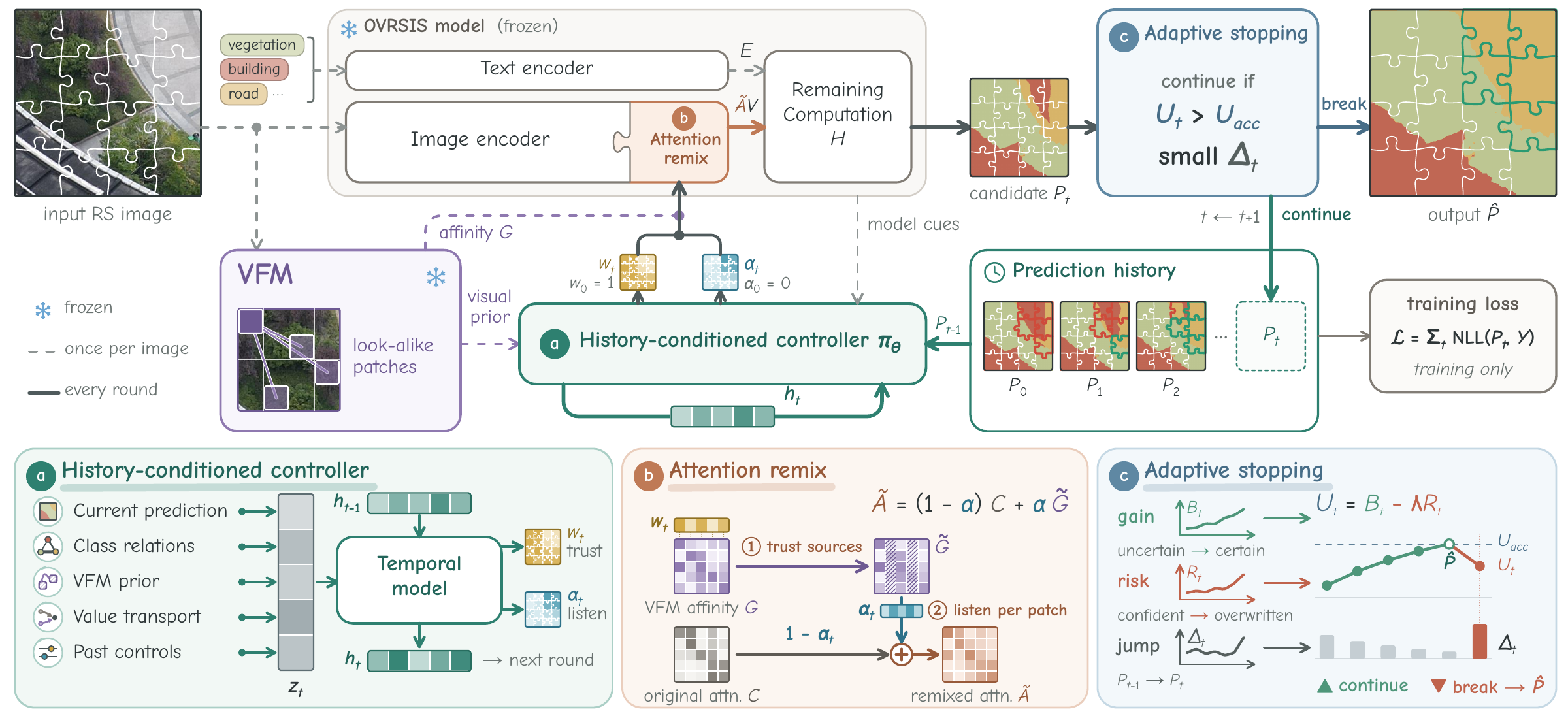}
\caption{Overall framework of Puzzle. Starting from the single-pass prediction, Puzzle combines prediction history with a visual-similarity prior to adjust patch-level attention and re-infer the frozen OVRSIS model. The history is updated across rounds, while adaptive stopping determines when to return the last accepted prediction. (a) History-conditioned controller; (b) attention remix; (c) adaptive stopping.}
\label{fig:framework}
\end{figure*}

Given an RS image divided into $N$ patches and $K$ candidate classes, a CLIP-based OVRSIS model predicts segmentation labels by matching visual features with the text embeddings $E$ of the class names. In the visual encoder, self-attention controls how information is exchanged among patches. Let $A$ and $V$ denote the attention weights and value features at the final attention layer, respectively. For a fixed input, we denote the remaining prediction process, including output projection, optional feature upsampling, and visual-text matching, by $\mathcal{H}$, and write the single-pass prediction as $P_0=\mathcal{H}(AV,E)$. We extract the patch-to-patch attention from $A$ and normalize each row to obtain the CLIP attention matrix $C\in\mathbb{R}^{N\times N}$, which describes how each patch attends to other patches.

While $C$ describes the model's attention among patches, we additionally use a frozen vision foundation model (VFM) to capture their visual similarity. Specifically, we extract VFM features aligned with the same patch grid and denote the $\ell_2$-normalized feature of patch $i$ by $f_i$. The visual affinity between patches is defined as:
\begin{equation}
	G_{ij}=\frac{\exp(f_i^\top f_j/\tau)}
	{\sum_{k=1}^{N}\exp(f_i^\top f_k/\tau)},
	\label{eq:visual_affinity}
\end{equation}
where $\tau>0$ is a temperature parameter. A larger $G_{ij}$ indicates greater visual similarity between patches $i$ and $j$, and we refer to patches with high affinity to patch $i$ as its look-alikes. Thus, $C$ represents the model's current attention, while $G$ provides a visual-similarity prior. Puzzle uses these two relations and the prediction history to adjust information exchange among patches during iterative re-inference.

\section{Method}
\label{sec:method}

\subsection{Overview}
\label{sec:overview}

Fig.~\ref{fig:framework} illustrates Puzzle, which performs iterative re-inference on an existing OVRSIS model with an attention remix inserted into its final layer. At round $t\geq1$, a history-conditioned controller $\pi_\theta$ uses the prediction history and a VFM's visual-similarity prior to estimate two controls for each patch $i$: source trust $w_{t,i}$, which determines its weight as an information source, and listening strength $\alpha_{t,i}$, which controls how much information it receives from visually similar patches. These controls adjust the attention weights in the visual encoder's final layer, allowing the frozen model to produce a new prediction $P_t$. At round 0, setting $w_0=1$ and $\alpha_0=0$ keeps the original attention and reproduces the single-pass prediction $P_0$. After each round, an adaptive stopping rule decides whether to continue or return the last accepted prediction. The controller is trained once and reused across models and datasets. We present the inference process in Algorithm \ref{alg:infer}.

\suppressfloats[t]
\begin{algorithm}[t]
    \caption{Inference procedure of \ours}
    \label{alg:infer}
    \small
    \begin{algorithmic}[1]
        \Require image $I$, class names, frozen OVRSIS model and VFM, controller $\pi_\theta$
        \State Compute and cache $C$, $V$, $G$ (Eq.~(\ref{eq:visual_affinity})) \Comment{once per image}
        \State $w_0\gets1$, $\alpha_0\gets0$, $h_0\gets0$ \Comment{initialize controls}
        \State $\tilde{A}_0\gets$ Eqs.~(\ref{eq:trust})--(\ref{eq:listen}); $P_0\gets\mathcal{H}(\tilde{A}_0V,E)$ \Comment{single pass}
        \State $\hat{P}\gets P_0$, $U_{\mathrm{acc}}\gets0$
        \For{$t=1,\dots,T_{\max}$}
        \State $(w_t,\alpha_t,h_t)\gets\pi_\theta(P_{t-1},w_{t-1},\alpha_{t-1},h_{t-1})$
        \State $\tilde{A}_t\gets$ Eqs.~(\ref{eq:trust})--(\ref{eq:listen}); $P_t\gets\mathcal{H}(\tilde{A}_tV,E)$
        \State compute $B_t$, $R_t$, $\Delta_t$, $U_t$ by Eqs.~(\ref{eq:gain})--(\ref{eq:xiaoyong})
        \If{$t>W$ and $P_t$ is rejected (Sec.~\ref{sec:stop})} \textbf{break} \EndIf
        \State $\hat{P}\gets P_t$, $U_{\mathrm{acc}}\gets U_t$ \Comment{accept $P_t$}
        \If{$P_t$ has converged} \textbf{break} \EndIf
        \EndFor
        \State \Return $\hat{P}$
    \end{algorithmic}
\end{algorithm}

\subsection{History-Conditioned Controller}
\label{sec:controller}

Identifying reliable references is challenging because visually similar patches may also be misclassified. Moreover, patches differ in how much information they need to receive, and these requirements may change across inference rounds. We therefore design a history-conditioned controller $\pi_\theta$ to estimate the source trust and listening strength of each patch from its prediction history and visual context. As shown in Fig.~\ref{fig:framework}(a), the controller takes five types of information as input: the current prediction, class relations, the VFM prior, value transport, and past controls. It projects them into compact evidence and accumulates this evidence across rounds with a temporal model.

The current prediction $P_{t-1}$ provides the class scores and uncertainty of each patch. For each class $k$, we use the centered logit, the probability $p_{t-1,ik}$, and the semantic support $(p_{t-1,i}S)_k$, where $S$ is the cosine-similarity matrix of the class text embeddings. A class-shared projector processes these features, and symmetric pooling across classes yields the prediction embedding. However, pooling discards which classes compete, and similar uncertainty may come from confusing related classes, such as low vegetation and tree, or unrelated ones. The class relations therefore combine $p_{t-1,i}$ with $S$ to let the controller distinguish the two cases. The VFM prior uses the visual affinity $G$ to aggregate the VFM features and class relations of the look-alikes, which indicates whether the look-alikes are reliable and whether patch $i$ can serve as a reference. The value transport consists of the local value $V_i$ and its two contexts $(CV)_i$ and $(GV)_i$, since adjusting attention changes how value features are aggregated. The past controls $w_{t-1,i}$ and $\alpha_{t-1,i}$, along with the round index $t$, allow subsequent decisions to account for earlier adjustments.

Each input is projected to a low-dimensional feature by a lightweight projector, and the projected features are concatenated into the fused evidence $z_{t,i}$. Since a single round provides only partial evidence about a patch, a temporal model maintains a memory $h_{t,i}$ for each patch and updates it as follows:
\begin{equation}
    h_{t,i}=\Lambda_{t,i}\odot h_{t-1,i}+\Gamma_{t,i}\,z_{t,i},
    \label{eq:state}
\end{equation}
where the decay $\Lambda_{t,i}$ and input map $\Gamma_{t,i}$ are computed from $z_{t,i}$ to control the retention of past evidence and the incorporation of current evidence, respectively, and $h_{0,i}=0$. Two linear heads then map $h_{t,i}$ to the source trust $w_{t,i}\in[w_{\min},1]$ and listening strength $\alpha_{t,i}\in[0,1]$. Architectural details are provided in the Appendix.

\subsection{Attention Remix}
\label{sec:remix}

Source trust and listening strength control two aspects of patch-level information exchange: how much each patch contributes as an information source and how much it receives from visually similar patches. As shown in Fig.~\ref{fig:framework}(b), Puzzle uses these controls to construct the remixed attention while keeping the OVRSIS model frozen.

We first reweight each column $j$ of the visual affinity $G$ by its source trust $w_{t,j}$ and normalize each row to obtain the trusted affinity:
\begin{equation}
	\tilde{G}_{t,ij}=\frac{G_{ij}\,w_{t,j}}{\sum_{k}G_{ik}\,w_{t,k}}.
	\label{eq:trust}
\end{equation}

\begin{figure}[t]
\centering
\includegraphics[width=\linewidth]{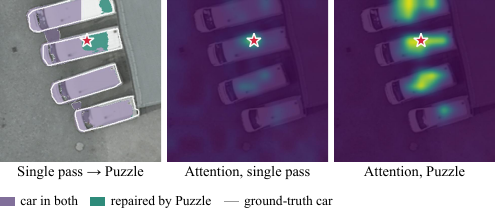}
\caption{Attention of a misclassified patch ($\bigstar$) before and after the remix (Potsdam, SegEarth-OV).} 
\label{fig:mechanism}
\end{figure}

This reduces the contribution of low-trust patches as information sources, even when they are visually similar to the receiving patch. We then use the listening strength to interpolate between the original CLIP attention $C_i$ and the trusted affinity $\tilde{G}_{t,i}$ for each patch $i$:
\begin{equation}
	\tilde{A}_{t,i}=(1-\alpha_{t,i})\,C_i+\alpha_{t,i}\,\tilde{G}_{t,i}.
	\label{eq:listen}
\end{equation}
When $\alpha_{t,i}=0$, patch $i$ retains its original attention; as $\alpha_{t,i}$ increases, more weight is assigned to the trusted affinity. The frozen OVRSIS model then re-infers with the remixed attention to obtain $P_t=\mathcal{H}(\tilde{A}_tV,E)$. As illustrated in Fig.~\ref{fig:mechanism}, the remixed attention shifts toward visually similar regions of the same class, correcting the misclassified patch in this example.

\subsection{Adaptive Stopping}
\label{sec:stop}

The benefit of re-inference varies across images and rounds. Stopping too early may leave misclassified regions uncorrected, while continuing too long may disturb regions that were already correctly classified. A manually selected number of rounds cannot accommodate these image-specific differences. Since ground-truth labels are unavailable during inference, selecting the best round for each image is also impractical. We therefore introduce an adaptive stopping rule that uses the predicted class distributions to decide whether to accept each new prediction and when to terminate.

Let $\bar{H}$ denote the normalized entropy and $u_i=\bar{H}(p_{0,i})$ the initial uncertainty of patch $i$. For each new prediction $P_t$, we compute three quantities:
\begin{align}
	B_t&=\mean_i[u_i\,(\bar{H}(p_{0,i})-\bar{H}(p_{t,i}))],\label{eq:gain}\\
	R_t&=\mean_i[(1-u_i)JS(p_{0,i},p_{t,i})/\log 2],\label{eq:risk}\\
	\Delta_t&=\mean_i[JS(p_{t-1,i},p_{t,i})/\log 2].\label{eq:jump}
\end{align}
where $B_t$ measures the entropy reduction in initially uncertain patches, $R_t$ measures deviations from the initial predictions in initially confident patches, and $\Delta_t$ measures the prediction change between consecutive rounds. Inspired by the average--risk trade-off in portfolio selection \cite{markowitz1952portfolio}, we define the risk-adjusted utility as:
\begin{equation}
	U_t=B_t-\lambda R_t.
	\label{eq:xiaoyong}
\end{equation}

The inter-round change $\Delta_t$ helps identify abrupt prediction shifts that may not be captured by $U_t$ alone. In the Puzzle, we reject $P_t$ if $\Delta_t$ is substantially larger than both $\Delta_{t-1}$ and $\Delta_{t-2}$, or if $U_t \leq U_{\text{acc}}$, where $U_{\text{acc}}$ denotes the utility of the last accepted prediction. Otherwise, $P_t$ is accepted, and iteration continues until the predictions converge or $T_{\max}$ rounds are reached.

\subsection{Training}
\label{sec:training}

To enable plug-and-play use with existing OVRSIS models, we keep the segmentation model and VFM frozen and train only the lightweight controller $\pi_\theta$, which contains approximately 0.12M parameters. Using the training split of a single dataset, we unroll $T_{\max}$ rounds of re-inference and minimize the pixel-wise negative log-likelihood averaged across rounds:
\begin{equation}
	\mathcal{L}=\frac{1}{T_{\max}}\sum_{t=1}^{T_{\max}}\text{NLL}(P_t,Y),
	\label{eq:loss}
\end{equation}
where $Y$ denotes the ground-truth segmentation labels.

After training, the controller remains frozen and can be directly applied across OVRSIS models and datasets without further optimization. Centering the logits removes sensitivity to uniform logit shifts, while the class-shared encoder and symmetric pooling make the prediction embedding invariant to class ordering. Moreover, the controller has no class-specific parameters, so its size does not depend on the number of classes $K$. These properties support transfer across different class taxonomies, while the use of class-level and VFM-derived cues reduces dependence on a particular segmentation model.

\subsection{Differences from Previous Studies}
The studies most closely related to Puzzle are ProxyCLIP \cite{lan2024proxyclip} and FSA \cite{chi2025fsa}. ProxyCLIP adjusts CLIP attention using VFM-derived visual similarity, while FSA feeds predictions back into the attention. However, both apply fixed attention adjustments within a single inference pass. In contrast, Puzzle performs iterative re-inference, where a lightweight controller combines the visual-similarity prior with prediction history to adjust attention across rounds. This allows regions corrected in earlier rounds to guide subsequent predictions, while an adaptive stopping rule determines when to terminate. Moreover, Puzzle is a plug-and-play framework that can be applied to existing OVRSIS models to further improve their segmentation performance without retraining them.

\section{Experiments}
\label{sec:exp}

\begin{table*}[t]
\centering
\caption{Comparison with SOTA methods. Multi-class: mIoU; single-class: IoU of the foreground class. Puzzle uses the controller trained on FLAIR. Baselines on iSAID and the single-class datasets are from~\cite{li2025segearthov,yang2026tmpa,niu2026reattnclip}.}
\label{tab:main}
\setlength{\tabcolsep}{1.1pt}
\footnotesize
\fitwidth{%
\begin{tabular}{@{\extracolsep{\fill}}l l cccccccc cc c@{}}
\toprule
& & \multicolumn{8}{c}{Multi-class semantic segmentation} & \multicolumn{2}{c}{Single-class extraction} & \\
\cmidrule(lr){3-10}\cmidrule(lr){11-12}
Method & Venue & OEM & LoveDA & iSAID & Potsdam & Vaihingen & UAVid & UDD5 & VDD & WHU$^{\text{Aerial}}$ & CHN6-CUG & Avg. \\
\midrule
CLIP~\cite{radford2021clip} & \venue{ICML'21} & 11.5 & 12.6 & 7.5 & 14.5 & 10.3 & 10.3 & 11.4 & 14.0 & 17.7 & 7.7 & 11.7 \\
MaskCLIP~\cite{zhou2022maskclip} & \venue{ECCV'22} & 24.7 & 26.0 & 14.5 & 31.7 & 24.7 & 28.4 & 34.2 & 33.2 & 29.8 & 28.1 & 27.5 \\
SCLIP~\cite{wang2024sclip} & \venue{ECCV'24} & 29.0 & 27.5 & 16.1 & 36.6 & 28.4 & 30.1 & 36.4 & 37.4 & 33.4 & 21.1 & 29.6 \\
GEM~\cite{bousselham2024gem} & \venue{CVPR'24} & 30.5 & 25.5 & 17.7 & 35.4 & 21.6 & 31.7 & 33.9 & 38.6 & 24.4 & 13.4 & 27.3 \\
ClearCLIP~\cite{lan2024clearclip} & \venue{ECCV'24} & 30.7 & 29.4 & 18.2 & 40.9 & 27.3 & 35.0 & 39.3 & 39.0 & 36.6 & 25.5 & 32.2 \\
ResCLIP~\cite{yang2025resclip} & \venue{CVPR'25} & 36.9 & 29.4 & 8.8 & 42.2 & 28.5 & 36.6 & 42.3 & 39.6 & -- & -- & \\
SegEarth-OV~\cite{li2025segearthov} & \venue{CVPR'25} & 39.7 & 36.9 & 21.7 & 47.1 & 29.1 & 42.5 & 50.6 & 45.3 & 49.2 & 35.4 & 39.8 \\
\midrule
TMPA~\cite{yang2026tmpa} (baseline) & \venue{CVPR'26} & 42.1 & 39.7 & 26.2 & 51.1 & 43.3 & 46.2 & 51.1 & 47.6 & 55.6 & 36.8 & 44.0 \\
TMPA + \ours & \venue{Ours} & \textbf{42.6}\gp{0.5} & \textbf{40.6}\gp{0.9} & \textbf{26.8}\gp{0.6} & \textbf{52.7}\gp{1.6} & \textbf{44.1}\gp{0.8} & \textbf{47.7}\gp{1.5} & \textbf{52.1}\gp{1.0} & \textbf{49.8}\gp{2.2} & \textbf{57.2}\gp{1.6} & \textbf{39.8}\gp{3.0} & \textbf{45.3}\gp{1.3} \\
\bottomrule
\end{tabular}}
\end{table*}

\def\GenTransferMain{%
\multirow{2}{*}{CLIP~\cite{radford2021clip}} & \LLoveDA & 20.0\gp{8.5} & \srcmark & 28.4\gp{13.9} & 20.8\gp{10.5} & 22.3\gp{12.0} & 24.9\gp{13.5} & 28.0\gp{14.0} & 24.1\gp{12.1} \\
 & \LOEM & \srcmark & 21.2\gp{8.6} & 24.3\gp{9.8} & 19.2\gp{8.9} & 17.6\gp{7.3} & 20.2\gp{8.8} & 22.8\gp{8.8} & 20.9\gp{8.7} \\
\midrule
\multirow{2}{*}{MaskCLIP~\cite{zhou2022maskclip}} & \LLoveDA & 26.0\gp{1.3} & \srcmark & 32.7\gp{1.0} & 26.1\gp{1.4} & 29.0\gp{0.6} & 36.1\gp{1.9} & 36.5\gp{3.3} & 31.1\gp{1.6} \\
 & \LOEM & \srcmark & 30.7\gp{4.7} & 34.4\gp{2.7} & 28.8\gp{4.1} & 29.3\gp{0.9} & 36.9\gp{2.7} & 36.5\gp{3.3} & 32.8\gp{3.1} \\
\midrule
\multirow{2}{*}{GEM~\cite{bousselham2024gem}} & \LLoveDA & 31.0\gp{0.5} & \srcmark & 38.9\gp{3.5} & 22.0\gp{0.4} & 33.9\gp{2.2} & 35.9\gp{2.0} & 39.3\gp{0.7} & 33.5\gp{1.6} \\
 & \LOEM & \srcmark & 28.4\gp{2.9} & 37.3\gp{1.9} & 22.7\gp{1.1} & 34.3\gp{2.6} & 36.6\gp{2.7} & 40.5\gp{1.9} & 33.3\gp{2.2} \\
\midrule
\multirow{2}{*}{SCLIP~\cite{wang2024sclip}} & \LLoveDA & 32.2\gp{3.2} & \srcmark & 39.1\gp{2.5} & 30.8\gp{2.4} & 32.0\gp{1.9} & 38.7\gp{2.3} & 41.3\gp{3.9} & 35.7\gp{2.7} \\
 & \LOEM & \srcmark & 30.4\gp{2.9} & 39.0\gp{2.4} & 31.0\gp{2.6} & 32.4\gp{2.3} & 38.4\gp{2.0} & 39.8\gp{2.4} & 35.2\gp{2.4} \\
\midrule
\multirow{2}{*}{ClearCLIP~\cite{lan2024clearclip}} & \LLoveDA & 32.6\gp{1.9} & \srcmark & 43.2\gp{2.3} & 29.6\gp{2.3} & 37.9\gp{2.9} & 44.4\gp{5.1} & 43.7\gp{4.7} & 38.6\gp{3.2} \\
 & \LOEM & \srcmark & 31.8\gp{2.4} & 42.5\gp{1.6} & 29.9\gp{2.6} & 36.7\gp{1.7} & 42.6\gp{3.3} & 41.1\gp{2.1} & 37.4\gp{2.3} \\
\midrule
\multirow{2}{*}{ResCLIP~\cite{yang2025resclip}} & \LLoveDA & 37.7\gp{0.8} & \srcmark & 43.0\gp{0.8} & 29.3\gp{0.8} & 38.6\gp{2.0} & 45.4\gp{3.1} & 42.2\gp{2.6} & 39.4\gp{1.7} \\
 & \LOEM & \srcmark & 32.7\gp{3.3} & 43.5\gp{1.3} & 31.3\gp{2.8} & 38.4\gp{1.8} & 43.6\gp{1.3} & 40.6\gp{1.0} & 38.4\gp{1.9} \\
\midrule
\multirow{2}{*}{SegEarth-OV~\cite{li2025segearthov}} & \LLoveDA & 39.9\gp{0.2} & \srcmark & 47.9\gp{0.8} & 30.1\gp{1.0} & 42.9\gp{0.4} & 52.2\gp{1.6} & 48.1\gp{2.8} & 43.5\gp{1.1} \\
 & \LOEM & \srcmark & 38.9\gp{2.0} & 48.2\gp{1.1} & 31.1\gp{2.0} & 42.9\gp{0.4} & 51.0\gp{0.4} & 45.8\gp{0.5} & 43.0\gp{1.1} \\
}

\begin{table*}[t]
\centering
\caption{Ablation results for controller training datasets across models (mIoU). Puzzle (X): controller trained on X.}
\label{tab:transfer}
\setlength{\tabcolsep}{3.2pt}
\footnotesize
\def\LLoveDA{+\,\ours{} (LoveDA)}
\def\LOEM{+\,\ours{} (OEM)}
\fitwidth{%
\begin{tabular}{@{\extracolsep{\fill}}l l cccccccc@{}}
\toprule
Model & Setting & OEM & LoveDA & Potsdam & Vaihingen & UAVid & UDD5 & VDD & Avg. \\
\midrule
\GenTransferMain
\bottomrule
\end{tabular}}
\end{table*}

\begin{figure*}[t]
\centering
\includegraphics[width=0.98\linewidth]{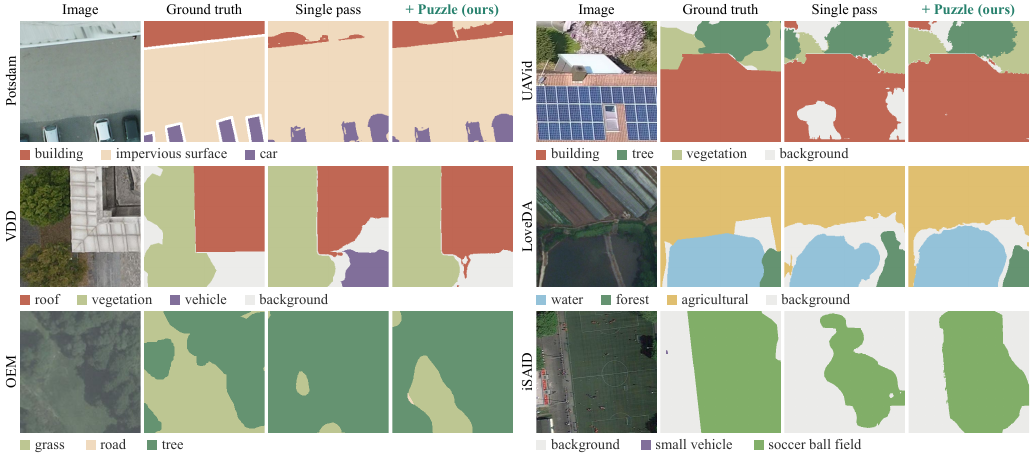}
\caption{Qualitative comparison.}
\label{fig:qualitative}
\end{figure*}

\subsection{Setup}
\label{sec:setup}

\noindent\textbf{Datasets.}
Following previous studies \cite{li2025segearthov,yang2026tmpa}, we evaluate Puzzle on eight multi-class semantic segmentation datasets: OpenEarthMap (OEM) \cite{xia2023openearthmap}, LoveDA \cite{wang2021loveda}, iSAID \cite{zamir2019isaid}, Potsdam and Vaihingen \cite{rottensteiner2012isprs}, UAVid \cite{lyu2020uavid}, UDD5 \cite{chen2018udd}, and VDD \cite{cai2023vdd}. We further evaluate building extraction on WHU$^{\text{Aerial}}$ \cite{ji2019whu} and road extraction on CHN6-CUG \cite{zhu2021chn6}. The input size is set to $448\times448$. Further details are provided in the Appendix.

\noindent\textbf{Compared Methods.}
We compare Puzzle with seven CLIP-based single-pass methods: CLIP~\cite{radford2021clip}, MaskCLIP~\cite{zhou2022maskclip}, SCLIP~\cite{wang2024sclip}, GEM~\cite{bousselham2024gem}, ClearCLIP~\cite{lan2024clearclip}, ResCLIP~\cite{yang2025resclip}, and SegEarth-OV~\cite{li2025segearthov}, as well as the test-time adaptation method TMPA~\cite{yang2026tmpa}. On the multi-class datasets except iSAID, we evaluate the seven single-pass methods using a unified codebase, while TMPA follows the authors' released implementation and settings. For iSAID and the two single-class datasets, we use baseline results reported in previous studies~\cite{li2025segearthov,yang2026tmpa,niu2026reattnclip}.

\noindent\textbf{Implementation details.}
Puzzle keeps the underlying segmentation model frozen and adopts \dinov{} ViT-L/16 pre-trained on satellite imagery~\cite{simeoni2025dinov3} as the VFM. For comparison with SOTA methods, we apply \ours to TMPA using a controller trained on FLAIR \cite{garioud2022flair}, ensuring that none of the evaluation datasets is used to train this controller. For the plug-and-play experiments, we train two separate controllers on LoveDA and OEM, respectively. We use average Intersection over Union (mIoU) as the evaluation metric. Further details are provided in the Appendix.

\subsection{Main Experiments}
\label{sec:main}\label{sec:plug}\label{sec:single}

\noindent\textbf{Comparison with SOTA methods.}
\Cref{tab:main} compares \ours with SOTA methods. Iterative re-inference updates no model parameters and can therefore be stacked on existing methods: we attach \ours to TMPA, the most accurate of them. TMPA calibrates the text embeddings of the classes at test time based on high-confidence visual features, whereas \ours adjusts the information flow among patches in the visual encoder; the two act on different parts of the model. With \ours, TMPA improves on all ten datasets and raises the average score from 44.0 to 45.3 (+1.3). On the eight multi-class datasets, the gains range from 0.5 to 2.2 mIoU. The largest improvement is on VDD, from 47.6 to 49.8 (+2.2), with further gains on Potsdam (51.1 to 52.7, +1.6) and UAVid (46.2 to 47.7, +1.5). The iSAID result also increases from 26.2 to 26.8 (+0.6). On single-class extraction, the IoU rises from 55.6 to 57.2 (+1.6) on WHU$^{\text{Aerial}}$ and from 36.8 to 39.8 (+3.0) on CHN6-CUG, both the best results in the table. Trained on multi-class data, the controller also improves the extraction of buildings and roads, showing that the gains of iterative re-inference generalize to single-class extraction.

\noindent\textbf{Plug-and-play.}
To evaluate the transferability of Puzzle, we train two controllers separately on LoveDA and OEM, which have different class taxonomies, and apply each to seven CLIP-based segmentation models without further training, as shown in Table \ref{tab:transfer}. Puzzle consistently improves the single-pass predictions across all evaluated settings, with average gains exceeding 3 mIoU for both controllers. The improvements are particularly pronounced for the original CLIP baseline but also extend to stronger segmentation models. Despite being trained on datasets with different taxonomies, the two controllers achieve comparable improvements, indicating that Puzzle transfers across different training datasets and segmentation models.

\begin{table}[!b]
\centering
\caption{Sensitivity to the number of remixed layers (mIoU).}
\label{tab:sens}
\setlength{\tabcolsep}{5pt}
\footnotesize
\begin{tabular}{@{}l ccc@{}}
\toprule
Remixed layers & SegEarth-OV & ClearCLIP & Avg. \\
\midrule
\multicolumn{4}{@{}l}{\textit{UDD5}}\\
Last layer & \textbf{52.16} & \textbf{44.41} & \textbf{48.28} \\
Last 2 layers & 48.51 & 24.23 & 36.37 \\
Last 3 layers & 46.06 & 24.93 & 35.50 \\
\midrule
\multicolumn{4}{@{}l}{\textit{VDD}}\\
Last layer & \textbf{48.05} & \textbf{43.70} & \textbf{45.88} \\
Last 2 layers & 33.16 & 27.32 & 30.24 \\
Last 3 layers & 23.30 & 20.53 & 21.91 \\
\bottomrule
\end{tabular}
\end{table}

\noindent\textbf{Qualitative results.}
Fig.~\ref{fig:qualitative} compares the segmentation results of SegEarth-OV with and without \ours. The refined predictions align more closely with the ground truth. In particular, \ours corrects misclassified building regions in Potsdam and UAVid, as well as roof regions mistakenly labeled as vehicles in VDD. The LoveDA example also shows improved separation of water, agricultural land, and forest, while regions already classified correctly remain largely unchanged.

\FloatBarrier

\subsection{Sensitivity to the Remixed Layers}
\label{sec:sens}

Puzzle remixes only the final attention layer. As a sensitivity analysis, we apply the same controls to the last $L\in\{1,2,3\}$ layers of SegEarth-OV and ClearCLIP on UDD5 and VDD, keeping the controller trained on LoveDA and the other settings unchanged. As shown in Table~\ref{tab:sens}, Puzzle is sensitive to this choice. Remixing only the final layer gives the best results in all cases, while remixing two or three layers falls below the single pass. Meanwhile, the risk $R_t$ in Eq.~(\ref{eq:risk}) increases with $L$: at the returned round, $R_t$ for $L=3$ is 3.0--12.1 times that for $L=1$. These results suggest that changes in earlier layers propagate through the subsequent layers and disturb initially confident predictions, so remixing the final layer alone is the most stable setting.

\subsection{Ablation Study and Analysis}
\label{sec:ablation}\label{sec:risk}

\begin{table}[!htbp]
\centering
\caption{Ablation results for the visual-similarity prior and historical information (mIoU).}
\label{tab:abl_info}
\setlength{\tabcolsep}{2.5pt}
\footnotesize
\fitwidth{%
\begin{tabular}{@{\extracolsep{\fill}}l ccccc@{}}
\toprule
Dataset & Prior & History & SegEarth-OV & ClearCLIP & Avg. \\
\midrule
\multirow{3}{*}{\textit{UDD5}} & $\times$ & $\times$ & 50.56 & 39.35 & 44.95 \\
& $\checkmark$ & $\times$ & 51.67 & 43.78 & 47.73\gp{2.78} \\
& $\checkmark$ & $\checkmark$ & \textbf{52.16} & \textbf{44.41} & \textbf{48.28}\gp{0.55} \\
\midrule
\multirow{3}{*}{\textit{VDD}} & $\times$ & $\times$ & 45.30 & 39.03 & 42.17 \\
& $\checkmark$ & $\times$ & 47.69 & 42.72 & 45.20\gp{3.03} \\
& $\checkmark$ & $\checkmark$ & \textbf{48.05} & \textbf{43.70} & \textbf{45.88}\gp{0.68} \\
\bottomrule
\end{tabular}}
\end{table}

We present the main ablation results here, with additional results provided in the Appendix.

Table \ref{tab:abl_info} examines the contributions of the visual prior and historical information. Starting from single-pass inference, we first enable iterative attention remix using the VFM prior, while resetting the controller's memory and past-control inputs at every round. This yields an average improvement of approximately 3 mIoU on both datasets. Restoring the historical information brings further gains, indicating that similarity-guided re-inference provides most of the improvement, while history helps refine subsequent adjustments.

\begin{table}[!htbp]
\centering
\caption{Ablation of the controls (mIoU). One control is fixed for all patches and rounds.}
\label{tab:abl_out}
\setlength{\tabcolsep}{5pt}
\footnotesize
\fitwidth{%
\begin{tabular}{@{\extracolsep{\fill}}l ccc@{}}
\toprule
Variant & SegEarth-OV & ClearCLIP & Avg. \\
\midrule
\multicolumn{4}{@{}l}{\textit{UDD5}}\\
Trust all ($w\equiv1$) & 51.48 & 43.67 & 47.57 \\
Uniform listening ($\alpha\equiv0.5$) & 51.63 & 43.45 & 47.54 \\
\ours & \textbf{52.16} & \textbf{44.41} & \textbf{48.28} \\
\midrule
\multicolumn{4}{@{}l}{\textit{VDD}}\\
Trust all ($w\equiv1$) & 45.52 & 41.67 & 43.60 \\
Uniform listening ($\alpha\equiv0.5$) & 48.00 & 42.95 & 45.48 \\
\ours & \textbf{48.05} & \textbf{43.70} & \textbf{45.88} \\
\bottomrule
\end{tabular}}
\end{table}

Table \ref{tab:abl_out} evaluates the two controls by replacing either source trust with $w_{t,i}=1$ or listening strength with $\alpha_{t,i}=0.5$ for all patches and rounds. Both variants consistently underperform Puzzle across the evaluated models and datasets. These results show the benefit of adapting reference contributions and information reception to individual patches and inference rounds, rather than assigning uniform weights.

\ifdefined\PuzzleDiagnosticsPlaced\else
\global\let\PuzzleDiagnosticsPlaced\relax
\begin{figure*}[t]
\centering
\includegraphics[width=\linewidth]{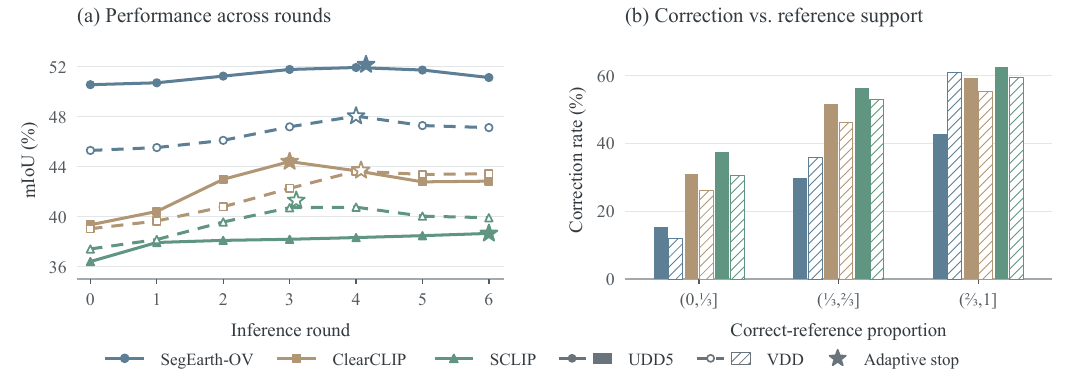}
\caption{(a) mIoU with six forced rounds; stars: adaptive stopping at its average returned round. (b) Correction rate of misclassified patches versus reference support, the share of correctly classified, same-class patches among their ten look-alikes.}
\label{fig:diagnostics}
\refstepcounter{subfigure}\label{fig:iter_rounds}
\refstepcounter{subfigure}\label{fig:support}
\end{figure*}
\fi

Fig.~\ref{fig:diagnostics}(a) compares adaptive stopping with fixed-round inference for SegEarth-OV, ClearCLIP and SCLIP on UDD5 and VDD. Without the stopping rule, the best round differs across models and datasets, ranging from the third to the sixth round. For example, ClearCLIP on UDD5 peaks at the third round and then loses 1.6 mIoU by the sixth round, whereas SCLIP on UDD5 still improves at the sixth round. In contrast, the adaptive stopping results, marked by stars, match or exceed the performance of every tested fixed-round setting, with a gain of up to 0.54 mIoU over the best one. This demonstrates the benefit of deciding when to stop rather than manually specifying a common number of rounds.

Finally, Fig.~\ref{fig:diagnostics}(b) examines how correction rates relate to the availability of correctly classified look-alikes. For each misclassified patch, we define reference support as the proportion of correctly classified, same-class patches among its ten most visually similar patches. The correction rate increases consistently with reference support for all three models on both datasets, rising from 12\%--37\% for support up to $1/3$ to 43\%--63\% for support above $2/3$. Patches without reference support are rarely corrected (2\%--11\%, not shown). This trend is consistent with the observation in Fig.~\ref{fig:evidence} and shows that patches with more correctly classified look-alikes are more likely to be corrected by Puzzle.

\FloatBarrier
\section{Related Work}
\label{sec:related}

\paragraph{Open-vocabulary segmentation.}
Existing open-vocabulary segmentation methods can be broadly grouped into three paradigms, as illustrated in Fig.~\ref{fig:teaser}. Single-pass methods adapt CLIP for dense prediction by modifying its self-attention~\cite{bousselham2024gem,hajimiri2025naclip,lan2024clearclip,wang2024sclip,yang2025resclip,zhou2022maskclip}. Other methods, including ProxyCLIP~\cite{lan2024proxyclip} and CASS~\cite{kim2025cass}, use VFM features to guide attention. For remote sensing (RS), related methods further adapt visual features, attention, or model weights~\cite{li2025farslip,li2025segearthov,liu2024remoteclip,niu2026reattnclip,wang2024skyscript,zhang2024georsclip}. Post-hoc refinement methods improve completed predictions using dense conditional random fields (CRFs)~\cite{krahenbuhl2011crf}, mask refinement~\cite{cheng2020cascadepsp,yuan2020segfix}, or score propagation over VFM-based patch graphs~\cite{chen2026coninfer,stojnic2025lposs,wang2026sdci}. Test-time adaptation (TTA) methods instead update model parameters or prompts using unsupervised objectives~\cite{niu2022eata,shu2022tpt,sun2020ttt,wang2021tent}, with applications to open-vocabulary segmentation~\cite{noori2025mlmp} and OVRSIS~\cite{yang2026tmpa}. Beyond CLIP-based approaches, recent work also explores SAM~3~\cite{carion2026sam3} for open-vocabulary segmentation. SegEarth-OV3~\cite{li2025segearthov3} combines its semantic and instance predictions and filters out absent classes using a presence score.

\paragraph{Iterative inference.}
Iterative inference has been explored in pose estimation~\cite{carreira2016ief}, recognition~\cite{zamir2017feedback}, optical flow~\cite{teed2020raft}, and text generation~\cite{madaan2023selfrefine}. In training-free segmentation, FSA~\cite{chi2025fsa} feeds prediction-derived affinity back into CLIP attention in a single step, CaR~\cite{sun2024car} iteratively prunes the queried class vocabulary, and iSeg~\cite{sun2026iseg} iteratively refines cross-attention maps using entropy-reduced self-attention. Post-hoc refinement can also be iterative~\cite{krahenbuhl2011crf,wang2026sdci}, but it updates prediction scores rather than re-running the segmentation model. In contrast, Puzzle performs iterative re-inference by combining prediction history with a visual-similarity prior to adjust patch-to-patch attention across rounds, while adaptively determining when to stop.

\section{Conclusion \& Discussion}
\label{sec:conclusion}

In this paper, we find through statistical analysis that many misclassified patches have visually similar, correctly classified patches of the same class within the same image. Motivated by this observation, we propose Puzzle, an iterative re-inference framework that combines prediction history with a visual-similarity prior to progressively adjust attention among patches while keeping the segmentation model frozen. With a controller trained only once, Puzzle provides plug-and-play refinement for different OVRSIS models without retraining them. Extensive experiments demonstrate consistent improvements across different OVRSIS models and datasets, supporting the transferability of Puzzle. 

Despite these improvements, Puzzle introduces additional inference overhead due to the computation of visual-similarity priors and repeated re-inference. Although adaptive stopping reduces unnecessary iterations, the computational cost remains higher than that of single-pass inference. Future work will explore lightweight alternatives to repeated downstream inference, aiming to retain the benefits of iterative correction while reducing computational overhead.\label{endofmaintext}

\nocite{bousselham2024gem,cai2023vdd,carion2026sam3,carreira2016ief,chen2018udd,chen2026coninfer,cheng2020cascadepsp,chi2025fsa,cho2014gru,garioud2022flair,gu2024mamba,gupta2019xbd,hajimiri2025naclip,ji2019whu,kim2025cass,krahenbuhl2011crf,lan2024clearclip,lan2024proxyclip,li2025farslip,li2025segearthov,li2025segearthov3,liu2024remoteclip,lyu2020uavid,madaan2023selfrefine,markowitz1952portfolio,niu2022eata,niu2026reattnclip,noori2025mlmp,radford2021clip,rottensteiner2012isprs,shu2022tpt,simeoni2025dinov3,stojnic2025lposs,sun2020ttt,sun2024car,sun2026iseg,teed2020raft,vaswani2017attention,wang2021loveda,wang2021tent,wang2024sclip,wang2024skyscript,wang2026sdci,xia2023openearthmap,yang2025resclip,yang2026tmpa,yuan2020segfix,zamir2017feedback,zamir2019isaid,zhang2024georsclip,zhou2022maskclip,zhu2021chn6}

\bibliographystyle{ieeenat_fullname}
\bibliography{main}

\clearpage
\appendix
\section*{Appendix}
\allowdisplaybreaks

\setcounter{table}{0}
\renewcommand{\thetable}{S\arabic{table}}
\renewcommand{\theHtable}{S\arabic{table}}
\setcounter{figure}{0}
\renewcommand{\thefigure}{S\arabic{figure}}
\renewcommand{\theHfigure}{S\arabic{figure}}
\setcounter{equation}{0}
\renewcommand{\theequation}{S\arabic{equation}}
\renewcommand{\theHequation}{S\arabic{equation}}
\setcounter{algorithm}{0}
\renewcommand{\thealgorithm}{S\arabic{algorithm}}
\providecommand{\theHalgorithm}{}
\renewcommand{\theHalgorithm}{S\arabic{algorithm}}

\noindent\ifdefined\BuildSupplement
References are listed at the end of this supplement, with the same numbering as in the main paper.
\else
This appendix shares the references of the main paper.
\fi Appendix~\ref{sec:s_theory} gives two properties of \ours; Appendix~\ref{sec:s_evidence} gives the statistics of correctly classified look-alikes of the same class (Fig.~\ref{fig:evidence} of the main paper); Appendix~\ref{sec:s_results} gives the per-class results; Appendix~\ref{sec:s_uncertainty} examines statistical uncertainty; Appendix~\ref{sec:s_qual} gives more qualitative results; Appendix~\ref{sec:s_ablation} compares different temporal models and retrieves the look-alikes with another VFM; Appendix~\ref{sec:s_impl} gives the implementation details. Unless stated otherwise, the results of the appendix use the controller trained on LoveDA in Tab.~\ref{tab:transfer}.

\section{Theoretical Properties}
\label{sec:s_theory}

This section gives two properties of \ours used in the main paper: the class-shared controller is invariant to class ordering and uniform logit shifts (Sec.~\ref{sec:controller} of the main paper), and the listening strength bounds how far a patch deviates from the CLIP attention (Sec.~\ref{sec:remix} of the main paper).

\paragraph{Class-dependent inputs.}
At controller round $t\ge1$, the current prediction is $P_{t-1}$. Omitting the subscript $t-1$ below, the prediction cue of patch $i$ consists of $K$ triplets
\begin{equation}
  x_{ik}=\big[(L_{ik}-\bar{L}_i)/s,\;p_{ik},\;(p_iS)_k\big],\qquad k=1,\dots,K,
  \label{eq:s_triplet}
\end{equation}
where $\bar{L}_i$ is the average logit of the patch, $S=EE^\top$ consists of the cosine similarities among the class names, and $s=10$ is a fixed scaling factor. The cue of class relations consists of the class-pair features $y_{ikl}=[p_{ik},p_{il},S_{kl}]$ and the single-class features $[p_{ik},(p_iS)_k]$. The network $\phi$, shared across all classes, encodes the triplets; the network $\psi$, shared across all class pairs and classes, encodes the relation features. The outputs of both are averaged over the class dimension and over the class pairs, both uniformly and with probability weights, the weights being $p_{ik}$ and $p_{ik}p_{il}$, respectively. The remaining inputs are either class-agnostic, such as the VFM features, the values and the past controls, or affinity-weighted averages of the relation summaries of the look-alikes.

\begin{proposition}[Class-agnostic decisions]
\label{prop:s_invariance}
\textup{(i)} In every round, the output of $\pi_\theta$ is unaffected by a permutation of the classes (applied jointly to $E$ and to the columns of the logits) and by adding an arbitrary constant to the logits of each patch. \textup{(ii)} Hence, if the class names are given in another order, \ours makes the same decisions in every round and stops at the same round, and the returned prediction only has its classes permuted accordingly. \textup{(iii)} The number of parameters of $\pi_\theta$ does not depend on $K$.
\end{proposition}

\noindent\textbf{Proof.} (i) Let $\sigma$ be a permutation of the classes. Then $S'_{kl}=S_{\sigma(k)\sigma(l)}$, $(p'_iS')_k=(p_iS)_{\sigma(k)}$, $x'_{ik}=x_{i\sigma(k)}$ and $y'_{ikl}=y_{i\sigma(k)\sigma(l)}$. A shared encoder followed by a (weighted) average over classes or class pairs is invariant to this relabeling. Adding the same constant to the logits of patch $i$ changes neither $L_{ik}-\bar{L}_i$ nor $p_i$. The remaining inputs are unaffected by construction. (ii) By induction over the rounds. If the decisions of the first $t-1$ rounds are the same, the remixed attention of Eqs.~(\ref{eq:trust}) and~(\ref{eq:listen}) of the main paper is also the same, since it depends only on the decisions, $C$ and $G$. The matching with the text embeddings then gives the same logits with permuted columns; by (i), the decisions of round $t$ are also the same. The scores of the stopping rule (Eqs.~(\ref{eq:gain})--(\ref{eq:jump}) of the main paper and \cref{eq:reject}) depend only on the entropies and Jensen--Shannon divergences of the class distribution of each patch and not on the order of the classes. (iii) The input dimensions of $\phi$ and $\psi$ are fixed, and pooling removes the dependence on $K$. \hfill$\square$

\smallskip
\Cref{prop:s_invariance} guarantees that a controller trained only on the classes of one dataset (\eg, the seven classes of LoveDA) can be used with a class vocabulary of any size and order, and that it responds only to class-agnostic quantities.

\begin{proposition}[Bounded deviation from the CLIP attention]
\label{prop:s_bounded}
For any round $t$, patch $i$ and any attention head with values $V$,
\begin{align}
  \big\|\tilde{A}_{t,i}-C_i\big\|_1&\le 2\,\alpha_{t,i},\label{eq:s_bound_attn}\\
  \big\|(\tilde{A}_tV)_i-(CV)_i\big\|_2&\le 2\,\alpha_{t,i}\max_j\|V_j\|_2.\label{eq:s_bound_out}
\end{align}
In particular, the probe of \cref{eq:s_probe} bounds the deviation of the first round by $2\alpha_{\mathrm{init}}$; and for a given reweighted affinity, the bounded step of \cref{eq:s_alpha} keeps the change of the attention row of a patch within $2\delta_\alpha$ in the $\ell_1$ norm.
\end{proposition}

\noindent\textbf{Proof.} By Eq.~(\ref{eq:listen}) of the main paper, $\tilde{A}_{t,i}-C_i=\alpha_{t,i}(\tilde{G}_{t,i}-C_i)$. Both $C_i$ and $\tilde{G}_{t,i}$ are probability vectors, whose $\ell_1$ distance is at most 2, which gives \cref{eq:s_bound_attn}. \Cref{eq:s_bound_out} follows from $\|\sum_j a_jV_j\|_2\le\|a\|_1\max_j\|V_j\|_2$. For two listening strengths $\alpha$ and $\alpha'$ and the same $\tilde{G}_{t,i}$, the difference of the two rows is $(\alpha-\alpha')(\tilde{G}_{t,i}-C_i)$, whose norm is at most $2|\alpha-\alpha'|\le2\delta_\alpha$.

\smallskip
For a native attention row with mass $m$, the bounds scale by $m$; the same argument applies to a remixed patch sub-row when its \cls entry is held fixed (Appendix~\ref{sec:s_hosts}). \Cref{prop:s_bounded} formalizes the statement of Sec.~\ref{sec:remix} of the main paper that a patch with a small listening strength stays close to its own attention; in addition, the probe limits how much the model is changed in the first round.

\section{Correctly Classified Look-Alikes of the Same Class}
\label{sec:s_evidence}

\paragraph{Protocol.}
We use the single pass of SegEarth-OV on the full evaluation sets of LoveDA, Potsdam and UDD5. Each image is divided by the patch grid of the controller, and the true and predicted classes of a patch are the majority labels of its pixels. Ignore-labeled target patches are excluded; the cached patch-majority convention is retained. For each misclassified patch $i$, we rank the other patches of the same image by the cosine similarity of their \dinov{} features, exclude the $5\times5$ neighborhood around $i$, and take the top-10 most similar patches whose cosine similarity is at least 0.8 as the look-alikes of $i$; a look-alike that belongs to the same true class as $i$ and is correctly classified by the single pass is called a \emph{correctly classified look-alike of the same class}. The ranking uses neither labels nor predictions, and the same neighborhood definition is used before and after re-inference. The example of Fig.~\ref{fig:evidence}(a,b) of the main paper is selected by the same rule: in Potsdam, we search for cars of which at least 70\% of the pixels are misclassified and whose look-alikes include at least two other correctly classified cars, and choose one of the qualifying examples.

\paragraph{Results.}
\Cref{tab:s_lookalike} gives the share of misclassified patches with at least one correctly classified look-alike of the same class, the shares of all retrieved look-alikes that have the same true class and that also predict it correctly, and the average number of correctly classified same-class look-alikes per misclassified patch. Of the 1,044,757 misclassified patches, 45.2\% have at least one correctly classified same-class look-alike. Across all retrieved look-alikes, 80.6\% have the same true class, while 21.0\% have the same true class and are correctly classified. Look-alikes often share the target's error, so a patch has to decide whom to trust (Sec.~\ref{sec:remix}).

\paragraph{After applying Puzzle.}
With the OEM-trained controller, the proportion in Fig.~\ref{fig:evidence}(c) decreases to 40.58\%, 38.89\% and 64.36\% on LoveDA, Potsdam and UDD5, respectively, and to 39.95\% overall. The overall reduction is 5.22 percentage points, indicating that \ours makes use of the correction potential in existing predictions.

\begin{table*}[t]
\centering
\caption{\textbf{Correctly classified look-alikes of the same class of misclassified patches} (single pass of SegEarth-OV). Coverage: share of misclassified patches with at least one such look-alike (95\% bootstrap interval, resampled by image for LoveDA and UDD5, and by original scene for Potsdam); Same: share of all look-alikes with the same true class as the target; Correct: share of all look-alikes with the same true class and a correct prediction; $n$: average number of the latter per misclassified patch.}
\label{tab:s_lookalike}
\setlength{\tabcolsep}{2.6pt}
\footnotesize
\begin{tabular*}{\linewidth}{@{\extracolsep{\fill}}l rr c c cc c@{}}
\toprule
Dataset & Images & Misclassified patches & Coverage (\%) & 95\% CI & Same (\%) & Correct (\%) & $n$ \\
\midrule
LoveDA & 1,669 & 577,735 & 44.7 & [43.8, 45.6] & 80.7 & 21.3 & 1.41 \\
Potsdam & 2,016 & 462,003 & 45.5 & [43.2, 47.7] & 80.6 & 20.6 & 1.62 \\
UDD5 & 40 & 5,019 & 66.1 & [61.4, 70.9] & 70.9 & 30.0 & 2.58 \\
\midrule
All & 3,725 & 1,044,757 & 45.2 & [44.0, 46.2] & 80.6 & 21.0 & 1.51 \\
\bottomrule
\end{tabular*}
\end{table*}

\section{Per-Class Results}
\label{sec:s_results}

\Cref{tab:s_perclass_a} gives the OEM per-class IoU for SegEarth-OV, ClearCLIP and SCLIP with the LoveDA-trained controller. The values are computed from the accumulated confusion matrices of the evaluations underlying Tab.~\ref{tab:transfer}. Single is the average single-pass IoU of these three models; the remaining columns give the changes produced by \ours. Class-level improvements and decreases are both retained.

\begin{table}[t]
\centering
\caption{\textbf{Per-class IoU on OEM} (full evaluation set). Single: single-pass IoU averaged over the three models; SE, CC, SC: change brought by \ours on SegEarth-OV, ClearCLIP and SCLIP, respectively; Avg.: average of the three changes.}
\label{tab:s_perclass_a}
\setlength{\tabcolsep}{2.4pt}
\footnotesize
\def\GenPerclassA{%
Background & 38.8 & \gain{+0.8} & \gain{+4.4} & \gain{+6.3} & \gain{+3.8} \\
Bareland & 14.7 & 0.0 & \loss{$-$0.8} & \gain{+0.5} & \loss{$-$0.1} \\
Grass & 22.3 & \gain{+0.4} & \gain{+5.2} & \gain{+8.6} & \gain{+4.7} \\
Pavement & 16.3 & \loss{$-$0.2} & \loss{$-$0.5} & \gain{+0.2} & \loss{$-$0.2} \\
Road & 21.5 & \gain{+0.4} & \gain{+2.7} & \gain{+4.3} & \gain{+2.5} \\
Tree & 47.5 & \gain{+0.3} & \gain{+2.4} & \gain{+2.5} & \gain{+1.7} \\
Water & 54.1 & \loss{$-$0.2} & \gain{+0.7} & \gain{+0.4} & \gain{+0.3} \\
Cropland & 44.5 & \loss{$-$0.1} & \loss{$-$1.3} & \gain{+2.5} & \gain{+0.4} \\
Building & 39.1 & \gain{+0.3} & \gain{+3.5} & \gain{+3.0} & \gain{+2.3} \\}

\begin{tabular*}{\linewidth}{@{\extracolsep{\fill}}l c cccc@{}}
\toprule
Class & Single & SE & CC & SC & Avg. \\
\midrule
\GenPerclassA
\bottomrule
\end{tabular*}
\end{table}

\section{Statistical Uncertainty Analysis}
\label{sec:s_uncertainty}

Each result of Tab.~\ref{tab:transfer} of the main paper comes from one trained controller and fixed evaluation sets. This section examines two sources of variation: the random seed of controller training (Appendix~\ref{sec:s_repeated}) and the sampling of the evaluation images (Appendix~\ref{sec:s_bootstrap}). Both analyses use the full evaluation sets.

\subsection{Repeated Runs}
\label{sec:s_repeated}

\paragraph{Protocol.}
We train the LoveDA-trained controller of Tab.~\ref{tab:transfer} three times with different random seeds, keeping the training data, the host model (SegEarth-OV) and all other settings unchanged, and evaluate each run on Potsdam and Vaihingen. Run 1 is the controller reported in the main paper.

\paragraph{Results.}
As shown in Tab.~\ref{tab:s_repeated}, all three runs improve over the single pass on both datasets, with average gains of +0.76 mIoU on Potsdam and +0.97 mIoU on Vaihingen. Across the runs, the mIoU differs by at most 0.11 on Potsdam and 0.05 on Vaihingen, and the standard deviations, 0.06 and 0.03, are below one tenth of the average gains. With $n=3$, we report these statistics as descriptive evidence of run-to-run stability rather than as a hypothesis test.

\begin{table}[t]
\centering
\caption{\textbf{Repeated runs} (SegEarth-OV, LoveDA-trained controller, full evaluation sets, mIoU). Run 1 is the controller of the main paper; Average $\pm$ Std: average and sample standard deviation over the three runs ($n=3$); $\Delta$: gain of the average over the single pass.}
\label{tab:s_repeated}
\footnotesize
\begin{tabular}{@{}l cc@{}}
\toprule
Setting & Potsdam & Vaihingen \\
\midrule
Single pass & 47.13 & 29.15 \\
\midrule
Run 1 & 47.88 & 30.12 \\
Run 2 & 47.94 & 30.15 \\
Run 3 & 47.83 & 30.10 \\
\midrule
Average $\pm$ Std & $47.89\pm0.06$ & $30.12\pm0.03$ \\
$\Delta$ & +0.76 & +0.97 \\
\bottomrule
\end{tabular}
\end{table}

\subsection{Paired Cluster Bootstrap}
\label{sec:s_bootstrap}

\paragraph{Protocol.}
Tiles cut from the same scene or video sequence are correlated, and resampling individual tiles would understate the uncertainty. We therefore resample the original scenes of Potsdam and Vaihingen and the video sequences of UAVid with replacement; these are the three datasets with complete paired predictions and a scene or sequence index for every tile. In each of 5,000 replicates, the confusion matrices of the single pass and of Puzzle are accumulated over the same resampled clusters, and the gain is the difference of the two mIoU values. The 95\% interval is given by the 2.5th and 97.5th percentiles of the replicates. The host is ClearCLIP with each of the two controllers of Tab.~\ref{tab:transfer}. Both controllers are trained with SegEarth-OV (Appendix~\ref{sec:s_train}), so these settings transfer the controller to an unseen host model and unseen datasets.

\paragraph{Results.}
Table~\ref{tab:s_bootstrap} gives the gains and their intervals; the point estimates are the ClearCLIP results of Tab.~\ref{tab:transfer} at higher precision. All six intervals exclude zero, so the gains are significant at the 5\% level under the paired cluster bootstrap; the smallest lower bound is 1.347 mIoU (OEM-trained controller, UAVid). These intervals describe the variation of the evaluation images for a fixed controller; the variation of controller training is examined in Appendix~\ref{sec:s_repeated}.

\begin{table}[t]
\centering
\caption{\textbf{Paired cluster bootstrap of the gains} (ClearCLIP, full evaluation sets, mIoU). Controller: training dataset of the controller; Single: single pass; $\Delta$: gain of Puzzle over the single pass; 95\% CI: percentile interval of $\Delta$ over 5,000 replicates, resampled by original scene for Potsdam and Vaihingen and by video sequence for UAVid.}
\label{tab:s_bootstrap}
\footnotesize
\setlength{\tabcolsep}{2pt}
\begin{tabular*}{\linewidth}{@{\extracolsep{\fill}}llrrrr@{}}
\toprule
Controller & Dataset & Single & Puzzle & $\Delta$ & 95\% CI \\
\midrule
\multirow{3}{*}{LoveDA} & Potsdam & 40.871 & 43.190 & +2.319 & [1.834, 2.855] \\
 & Vaihingen & 27.302 & 29.556 & +2.254 & [1.609, 2.985] \\
 & UAVid & 35.035 & 37.903 & +2.868 & [2.262, 3.596] \\
\midrule
\multirow{3}{*}{OEM} & Potsdam & 40.871 & 42.462 & +1.591 & [1.406, 1.793] \\
 & Vaihingen & 27.303 & 29.889 & +2.586 & [2.068, 3.039] \\
 & UAVid & 35.038 & 36.707 & +1.669 & [1.347, 1.973] \\
\bottomrule
\end{tabular*}
\end{table}

\section{More Qualitative Results}
\label{sec:s_qual}

\begin{figure*}[t]
\centering
\includegraphics[width=0.7\linewidth]{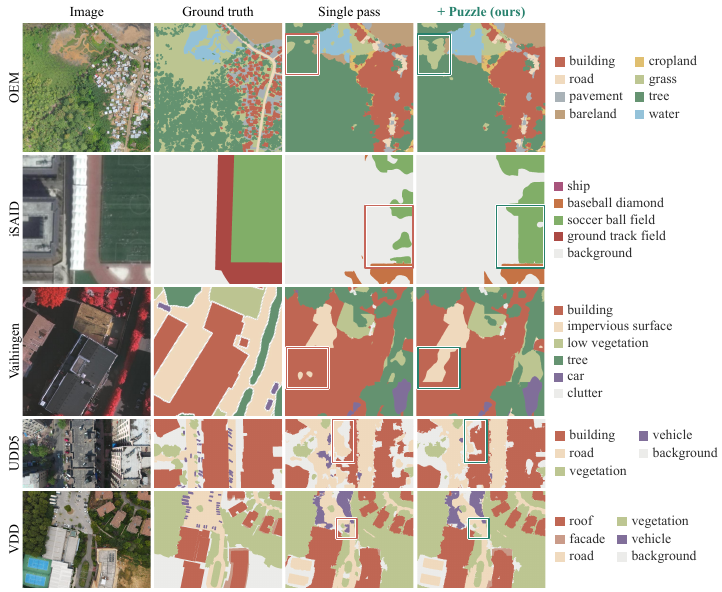}
\caption{\textbf{More qualitative results} (SegEarth-OV): one example each from OEM, iSAID, Vaihingen, UDD5 and VDD. Boxes mark the regions that the single pass misclassifies (terracotta) and \ours classifies correctly (teal).}
\label{fig:more_qual}
\end{figure*}

Fig.~\ref{fig:more_qual} shows more segmentation results of SegEarth-OV before and after attaching \ours, one example each from OEM, iSAID, Vaihingen, UDD5 and VDD; the images do not overlap with those of Fig.~\ref{fig:qualitative} of the main paper.

\section{Additional Ablations}
\label{sec:s_ablation}

\subsection{Temporal Model}
\label{sec:s_temporal}

The controller maintains a memory across rounds for every patch. \Cref{tab:s_temporal} compares, on Potsdam, the selective state-space model (Appendix~\ref{sec:s_ssm}) with a GRU~\cite{cho2014gru} of the same width and a causal Transformer~\cite{vaswani2017attention} along the rounds. The selective state-space model achieves the highest mIoU under both inference strategies. Adaptive stopping improves all three temporal models over three fixed rounds.

\begin{table}[t]
\centering
\caption{\textbf{Temporal model} (SegEarth-OV, Potsdam, mIoU).}
\label{tab:s_temporal}
\setlength{\tabcolsep}{8pt}
\footnotesize
\begin{tabular*}{\linewidth}{@{\extracolsep{\fill}}l cc@{}}
\toprule
Temporal model & \shortstack{Three fixed\\rounds} & \shortstack{Adaptive\\stopping} \\
\midrule
Selective state-space model (ours) & \textbf{46.58} & \textbf{47.88} \\
GRU & 46.28 & 47.80 \\
Causal Transformer & 46.34 & 47.49 \\
\bottomrule
\end{tabular*}
\end{table}

\subsection{Retrieving the Look-Alikes with Another VFM}
\label{sec:s_vfm}

We replace DINOv3 with the image encoder of SAM~3~\cite{carion2026sam3} and use a separately trained controller. As shown in \cref{tab:s_vfm}, both VFMs improve over the single pass in all six settings. DINOv3 performs better in five of the six settings, while the gains with SAM~3 indicate that \ours can also use visual similarity from another VFM.

\begin{table}[t]
\centering
\caption{\textbf{Source of the look-alikes} (mIoU). Each VFM uses its own controller.}
\label{tab:s_vfm}
\setlength{\tabcolsep}{1.8pt}
\footnotesize
\begin{tabular*}{\linewidth}{@{\extracolsep{\fill}}l l ccc@{}}
\toprule
Model & Dataset & Single pass & DINOv3 & SAM 3 \\
\midrule
\multirow{2}{*}{SegEarth-OV} & Potsdam & 47.13 & \textbf{47.88} & 47.27 \\
 & Vaihingen & 29.15 & 30.12 & \textbf{30.22} \\
\midrule
\multirow{2}{*}{ClearCLIP} & Potsdam & 40.87 & \textbf{43.19} & 42.01 \\
 & Vaihingen & 27.30 & \textbf{29.56} & 29.26 \\
\midrule
\multirow{2}{*}{SCLIP} & Potsdam & 36.62 & \textbf{39.07} & 38.42 \\
 & Vaihingen & 28.37 & \textbf{30.78} & 30.65 \\
\bottomrule
\end{tabular*}
\end{table}

\section{Implementation Details}
\label{sec:s_impl}

\subsection{Notation and Procedure}
\label{sec:s_procedure}

\Cref{tab:s_notation} summarizes the notation of this paper; the inference procedure of \ours is given in Algorithm~\ref{alg:infer} of the main paper.

\begin{table}[t]
\centering
\caption{\textbf{Notation.}}
\label{tab:s_notation}
\setlength{\tabcolsep}{3pt}
\footnotesize
\fitwidth{%
\begin{tabular}{@{\extracolsep{\fill}}l l@{}}
\toprule
Symbol & Meaning \\
\midrule
$I$, $K$, $N$ & image, number of classes, number of patches \\
$E$ & $\ell_2$-normalized class embeddings \\
$A$, $V$ & attention and values of the last layer \\
$C$ & row-normalized patch-to-patch attention \\
$\mathcal{H}(\cdot,E)$ & remaining prediction with class embeddings \\
$P_t$, $L$, $p_i$ & prediction of round $t$, patch logits and probabilities \\
$F$, $G$, $\tau$ & VFM features, affinity, temperature \\
$\alpha_{t,i}$, $w_{t,i}$ & listening strength, source trust \\
$\alpha_{\mathrm{init}}$, $\delta_\alpha$, $\delta_w$ & probe cap and control step caps \\
$w_{\min}$ & minimum source trust \\
$z_{t,i}$, $h_{t,i}$ & fused evidence, memory \\
$B_t$, $R_t$, $\Delta_t$, $U_t$ & gain, risk, jump, utility \\
$\lambda$, $\kappa$, $\epsilon$, $\epsilon'$ & risk weight and stopping thresholds \\
$W$, $T_{\max}$ & warm-up rounds and maximum round budget \\
\bottomrule
\end{tabular}}
\end{table}

\paragraph{Rejection criterion.}
After $W$ warm-up rounds, let $U_{\mathrm{acc}}$ be the utility of the last accepted round. As soon as the new prediction $P_t$ satisfies
\begin{equation}
\begin{aligned}
  &\Delta_t>\max\{\kappa\max(\Delta_{t-1},\Delta_{t-2}),\,\epsilon\}\\
  &\text{or}\quad U_t\le U_{\mathrm{acc}}+\epsilon',
\end{aligned}
  \label{eq:reject}
\end{equation}
it is rejected, and the last accepted prediction is returned. Here, the jump $\Delta_t$ and the utility $U_t$ are defined in Sec.~\ref{sec:stop} of the main paper, $\kappa$ is the factor of the jump relative to the two previous rounds, and $\epsilon$ and $\epsilon'$ are tolerances.

\subsection{Attaching \ours to CLIP-Based Models}
\label{sec:s_hosts}

The host model and VFM remain frozen. For $448\times448$ inputs, the controller uses a $28\times28$ grid, and DINOv3 features are average-pooled onto this grid. Sliding-window hosts use $224\times224$ crops with stride 112; GEM retains its native full-image path. Affinities and controls are aligned to the local patches before the host merges the window outputs. \Cref{tab:s_hosts} summarizes the native components retained by each host. TMPA supplies its calibrated text embeddings $E$, while \ours adjusts the final visual attention.

For a native token row with total attention mass $m_i$, a full-row remix can be written as
\begin{equation}
\tilde A_{t,i\cdot}=m_i\big((1-\alpha_{t,i})\bar C_{i\cdot}
+\alpha_{t,i}[0,\tilde G_{t,i\cdot}]\big),
\label{eq:s_mix}
\end{equation}
where $\bar C_{i\cdot}$ is the native row divided by $m_i$. The \cls query row is unchanged. If the host retains patch-to-\cls attention separately, the interpolation acts only on the patch sub-row and preserves its mass; if that entry is removed, the remaining patch row is normalized before remixing. Setting $\alpha_{t,i}=0$ recovers the native attention. Cached values are reused, and the remaining host-specific prediction process gives $P_t=\mathcal H(\tilde A_tV,E)$, as in Sec.~\ref{sec:remix}.

\begin{table}[t]
\centering
\caption{\textbf{Host-specific components retained by Puzzle.} All listed hosts use ViT-B/16 CLIP encoders. Softmax normalization and decoding follow each native implementation.}
\label{tab:s_hosts}
\setlength{\tabcolsep}{2.4pt}
\footnotesize
\begin{tabular*}{\linewidth}{@{\extracolsep{\fill}}l p{178pt}@{}}
\toprule
Host & Native components retained \\
\midrule
CLIP & $qk^\top$ attention, residual and FFN \\
MaskCLIP & identity attention, residual and FFN \\
GEM & iterative self-self attention and native decoder \\
SCLIP & sum of $qq^\top$ and $kk^\top$ softmaxes \\
ClearCLIP & $qq^\top$ softmax without final residual or FFN \\
ResCLIP & native RCS and SFR attention/refinement \\
SegEarth-OV & sum of $qq^\top$, $kk^\top$ and $vv^\top$ softmaxes \\
TMPA & native visual inference and calibrated text embeddings \\
\bottomrule
\end{tabular*}
\end{table}

\subsection{Controller Architecture}
\label{sec:s_arch}

\Cref{tab:s_params} lists the controller components shared across patches. The prediction encoder pools 32-dimensional class features and projects the result to a 64-dimensional residual branch. The fusion input combines eight 32-dimensional features: the relation summary and its $G$-weighted context, the VFM feature and its $G$-weighted context, and the projected values $V$, $CV$, $GV$ and their context difference. Together with the two previous controls and round index, they form a 259-dimensional input. The fusion output and prediction residual give $z_{t,i}\in\mathbb R^{64}$. The active controller has 118,754 parameters; the TMPA checkpoint has 123,011 parameters including an auxiliary readout. Inference uses the stopping rule of Sec.~\ref{sec:stop}.

\begin{table}[t]
\centering
\caption{\textbf{Parameters of the controller} (controller trained with SegEarth-OV).}
\label{tab:s_params}
\setlength{\tabcolsep}{1.8pt}
\footnotesize
\fitwidth{%
\begin{tabular}{@{\extracolsep{\fill}}l l r@{}}
\toprule
Component & Shape & Params \\
\midrule
Prediction encoder $\phi$ & $3\to32\to32$, pooling, $64\to32$ & 3,264 \\
\quad residual projection & $32\to64$ & 2,048 \\
Relation encoder $\psi$ & class pairs and classes, $128\to32$ & 6,560 \\
VFM feature projection & $1024\to32$ & 32,800 \\
Value projection & $768\to32$, LayerNorm & 24,672 \\
Fusion layer & $259\to64$, SiLU & 16,640 \\
Temporal model & width 64, inner 128, state 16 & 32,640 \\
Decision heads & $64\to2$ & 130 \\
\midrule
Total & & 118,754 \\
\bottomrule
\end{tabular}}
\end{table}

\subsection{Temporal Model}
\label{sec:s_ssm}

The temporal model is a selective state-space model in the style of Mamba~\cite{gu2024mamba}. It runs independently for every patch along the rounds and is reset at the start of every image. For the fused input $z_t\in\mathbb{R}^{64}$ of a patch (patch index omitted), it computes
\begin{align}
  (u_t,g_t)&=W_{\mathrm{in}}\operatorname{LN}(z_t),\label{eq:s_in}\\
  \hat{u}_t&=\operatorname{SiLU}\big(\textstyle\sum_{\tau=0}^{2}c_\tau\odot u_{t-\tau}+c_b\big),\label{eq:s_conv}\\
  (\rho_t,b_t,c_t)&=W_x\hat{u}_t,\label{eq:s_proj}\\
  \eta_t&=\operatorname{softplus}(W_\eta\rho_t+\beta_\eta),\label{eq:s_rate}\\
  s_t&=\exp(-\eta_t\odot a)\odot s_{t-1}+(\eta_t\odot\hat{u}_t)\,b_t^{\top},\label{eq:s_ssm}\\
  y_t&=(s_tc_t+D\odot\hat{u}_t)\odot\operatorname{SiLU}(g_t),\label{eq:s_read}\\
  h_t&=z_t+W_{\mathrm{out}}\,y_t,\label{eq:s_out}
\end{align}
with inner width 128, internal state $s_t\in\mathbb R^{128\times16}$, a causal convolution of width 3, positive rates $a=\exp(A_{\log})$ and skip parameter $D$. The state and convolution history are reset for each image. At round $t$, $z_t$ is computed from $P_{t-1}$; the 64-dimensional readout $h_t$ is passed to the decision heads, consistent with the round indexing of the main paper.

\paragraph{Decision heads.}
Two linear output heads map $h_t$ to the raw outputs $g^\alpha_t$ and $g^w_t$, which are converted into the controls by
\begin{align}
  \alpha_1&=\alpha_{\mathrm{init}}\,\sigma\big(g^\alpha_1\big),\qquad w_1=1,\label{eq:s_probe}\\
  \alpha_t&=\clip\big(\alpha_{t-1}+\delta_\alpha\tanh g^\alpha_t,\;0,\;1\big),\label{eq:s_alpha}\\
  w_t&=\clip\big(w_{t-1}+\delta_w\tanh g^w_t,\;w_{\min},\;1\big),\label{eq:s_w}
\end{align}
for $t\ge2$, with $w_{\min}=0.05$; training uses $\alpha_{\mathrm{init}}=0.2$ and $\delta_\alpha=\delta_w=0.3$. The round-0 controls remain $\alpha_0=0$ and $w_0=1$. \Cref{eq:s_probe} is the probe of the first round, and \cref{eq:s_alpha,eq:s_w} are the bounded steps of later rounds.

\subsection{Training}
\label{sec:s_train}

\paragraph{Controllers of Tab.~\ref{tab:transfer}.}
The controller of the LoveDA group is trained with SegEarth-OV on the LoveDA training split, with the CLIP model, the text encoder and \dinov{} frozen. Only the controller is optimized, using the pixel-wise negative log-likelihood averaged over the unrolled re-inference rounds in Eq.~(\ref{eq:loss}) of the main paper. The OEM controller is trained separately with SegEarth-OV on 256 OEM training images. Neither controller is further optimized when it is transferred to another host model or dataset.

\paragraph{Controller of Tab.~\ref{tab:main}.}
The controller attached to TMPA is trained in a single stage from random initialization, with TMPA, \dinov{} and the feature upsampler frozen. The training set consists of 2,000 FLAIR images from 32 French departments. Training unrolls six rounds ($\alpha_{\mathrm{init}}=0.2$, $\delta_\alpha=0.3$) and runs for three epochs (750 updates) with the AdamW optimizer, a learning rate of $10^{-4}$, a weight decay of 0.01 and a batch size of 8 on eight RTX 3090 GPUs; the checkpoint at the end of the third epoch was fixed before any dataset was evaluated.

\subsection{Datasets and Evaluation Protocol}
\label{sec:s_data}

For the eight multi-class datasets, we follow the evaluation sets and class names listed in \cref{tab:s_data}, following SegEarth-OV~\cite{li2025segearthov}, and compute mIoU from the confusion matrix accumulated over each dataset. The iSAID baselines are reported results, as stated in Sec.~\ref{sec:setup}. The ground truth of UAVid is taken from a public mirror of the dataset and converted with the official rules. The single-class extraction datasets WHU$^{\text{Aerial}}$ and CHN6-CUG follow the setting of TMPA~\cite{yang2026tmpa}, with $448\times448$ inputs and the IoU of the foreground class, on 1,036 and 903 evaluation images, respectively. The TMPA controller is trained on FLAIR, rather than any of these ten evaluation datasets, as stated in Sec.~\ref{sec:setup}. Tabs.~\ref{tab:sens}, \ref{tab:abl_info} and~\ref{tab:abl_out} use the 40-image UDD5 and VDD evaluation manifests.

\begin{table}[t]
\centering
\caption{\textbf{Evaluation sets} of the SegEarth-OV protocol.}
\label{tab:s_data}
\setlength{\tabcolsep}{3pt}
\footnotesize
\begin{tabular*}{\linewidth}{@{\extracolsep{\fill}}l l r@{}}
\toprule
Dataset & Data type & Images \\
\midrule
OpenEarthMap~\cite{xia2023openearthmap} & satellite / aerial, global & 384 \\
LoveDA~\cite{wang2021loveda} & satellite, urban and rural & 1,669 \\
iSAID~\cite{zamir2019isaid} & aerial, instance-level & 11,644 \\
Potsdam~\cite{rottensteiner2012isprs} & aerial, urban & 2,016 \\
Vaihingen~\cite{rottensteiner2012isprs} & aerial, urban & 398 \\
UAVid~\cite{lyu2020uavid} & UAV, oblique & 1,020 \\
UDD5~\cite{chen2018udd} & UAV, nadir & 40 \\
VDD~\cite{cai2023vdd} & UAV, nadir & 40 \\
\bottomrule
\end{tabular*}
\end{table}

\end{document}